\documentclass[letterpaper, 10 pt, conference]{ieeeconf}  

\IEEEoverridecommandlockouts                              

\usepackage{booktabs}
\usepackage{xcolor,colortbl}
\usepackage{graphicx}
\usepackage{multirow}
\usepackage[
    colorlinks=true,
    linkcolor=red,
    citecolor=green,
    urlcolor=magenta
]{hyperref}
\usepackage{makecell}
\let\labelindent\relax
\usepackage{enumitem}
\usepackage{amsmath}
\usepackage{algorithmic}
\usepackage{amsfonts}   
\usepackage{amssymb}
\usepackage{marvosym}

\title{\LARGE \bf
HeRO: Hierarchical 3D Semantic Representation for \\Pose-aware Object Manipulation
}

\author{Chongyang Xu$^{1,3*}$, Shen Cheng$^{3\dagger}$, Haipeng Li$^{2,3\dagger}$, Haoqiang Fan$^{3}$, Ziliang Feng$^{1}$ and Shuaicheng Liu$^{2}$\textsuperscript{\Letter}
\thanks{$^{*}$ This work was done during Chongyang Xu's internship at Dexmal.}
\thanks{$^{\dagger}$ Corresponding Project Leaders.}
\thanks{$^{1}$ College of Computer Science, Sichuan University,
$^{2}$ School of Information and Communication Engineering, University of Electronic Science and Technology of China,
$^{3}$ Dexmal.}
\thanks{\Letter \hspace{0.2em} Corresponding Author: liushuaicheng@uestc.edu.cn.}
}

\begin{document}

\newcommand{\cs}[1]{{\color{red}{#1}}}
\newcommand{\xcy}[1]{{\color{blue}{#1}}}

\newcommand{\figref}[1]{Fig.\ref{#1}}
\newcommand{\tabref}[1]{Tab.\ref{#1}}
\newcommand{\equref}[1]{Eqn.(\ref{#1})}
\newcommand{\secref}[1]{Sec.\ref{#1}}
\newcommand{\supref}[1]{\textbf{Suppl.~Sec.\ref{#1}}}

\newcommand{\best}[1]{\textbf{#1}}
\newcommand{\second}[1]{\uline{#1}}
\newcommand{\third}[1]{\textit{#1}}

\maketitle
\thispagestyle{empty}
\pagestyle{empty}

\begin{abstract}
Imitation learning for robotic manipulation has progressed from 2D image policies to 3D representations that explicitly encode geometry. Yet purely geometric policies often lack explicit \textit{part-level} semantics, which are critical for pose-aware manipulation (e.g., distinguishing a shoe’s  ``toe" from ``heel"). In this paper, we present \textbf{HeRO}, a diffusion-based policy that couples geometry and semantics via hierarchical \textit{semantic fields}. HeRO employs dense semantics lifting to fuse discriminative, geometry-sensitive features from DINOv2 with the smooth, globally coherent correspondences from Stable Diffusion, yielding dense features that are both fine-grained and spatially consistent. These features are processed and partitioned to construct a global field and a set of local fields. A hierarchical conditioning module conditions the generative denoiser on global and local fields using permutation-invariant network architecture, thereby avoiding order-sensitive bias and producing a coherent control policy for pose-aware manipulation. In various tests, HeRO establishes a new state-of-the-art, improving success on \textit{Place Dual Shoes} by \textbf{12.3\%} and averaging \textbf{6.5\%} gains across six challenging pose-aware tasks. Code is available at \url{https://github.com/Chongyang-99/HeRO}.
\end{abstract}
\section{INTRODUCTION}
Imitation learning for robotic manipulation~\cite{goyal2024rvt, zhao2023learning, wang2025oneshot, wu2025learning, dai2025racer} has advanced significantly, with policies evolving from 2D image-based methods~\cite{chi2023diffusion, liu_rdt-1b_2025} to 3D representations~\cite{ze20243d, ke_3d_2024, wilcox2025adapt3r, wang2024gendp, zhang2025flowpolicy, chen2025g3flow, jia_lift3d_2024}. While early image-based approaches achieve notable success, they often falter in tasks requiring precise spatial reasoning, as cameras inherently flatten the 3D world and lose crucial geometric information. To address this limitation, 3D imitation learning has emerged, leveraging representations like point clouds or voxels to explicitly model geometry.
A pioneering example is the 3D Diffusion Policy (DP3)~\cite{ze20243d}, which conditions its action generation on a compact 3D visual representation. By processing sparse point clouds, this approach can better capture spatial relationships for precise control, leading to improved performance.

\begin{figure}[tpb]
   \centering
   \includegraphics[width=\linewidth]{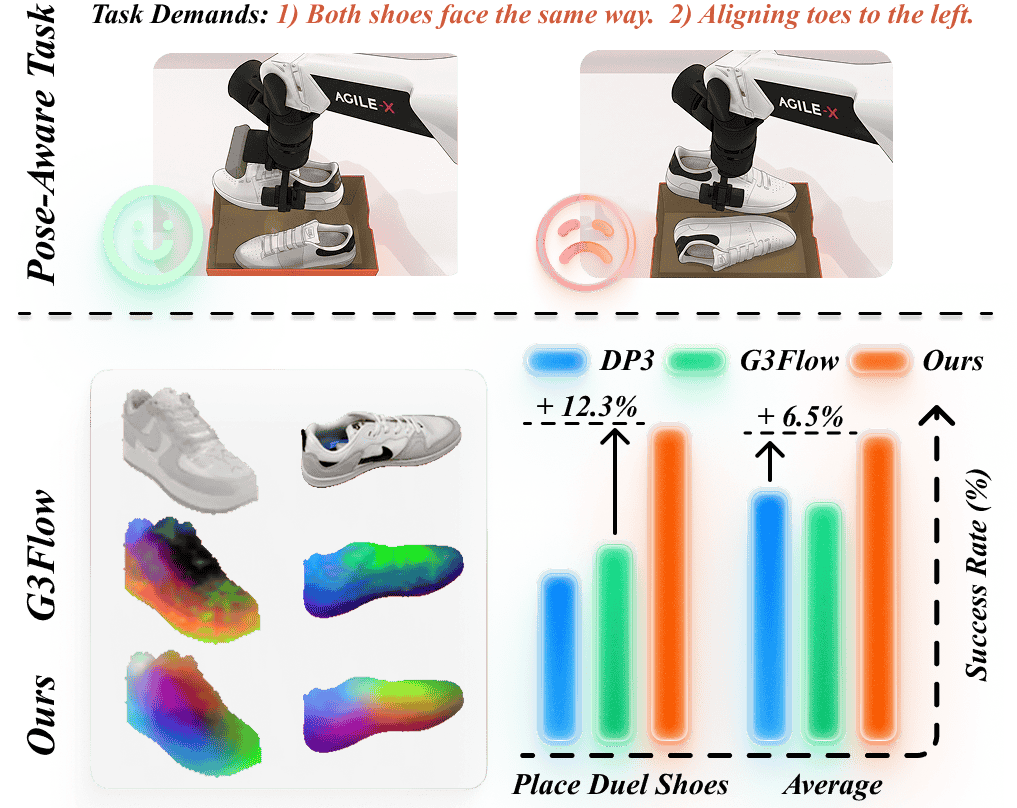}
   \vspace{-1em}
   \caption{\textbf{Pose-Aware Manipulation with Semantic Understanding.}
   \textbf{Top:} Many manipulation tasks are pose-aware (e.g., placing shoes with toes aligned left), demanding semantic part perception from policies.
   \textbf{Bottom Left:} Our dense semantic fields are smoother and more consistent than the baseline G3Flow~\cite{chen2025g3flow}.
   \textbf{Bottom Right:} Our method achieves a 12.3\% higher success rate on the dual shoe place task and a 6.5\% higher average success rate of 6 challenging tasks.}
   \label{fig:teaser}
   \vspace{-2em}
\end{figure}

However, despite their geometric strengths, these 3D methods often lack explicit semantic understanding. This limitation becomes a critical bottleneck in \textit{pose-aware} manipulation scenarios, where success hinges on identifying and reasoning about specific object parts. For instance, a task like placing shoes, as illustrated in Fig.~\ref{fig:teaser} (top), requires more than just successfully moving them to a location; it demands precise alignment based on functional parts like the \textit{``toe"} and \textit{``heel"}. Methods that rely purely on geometry cannot disambiguate between these semantically distinct parts, often leading to task failure.

\begin{figure}[tbp]
   \centering
   \includegraphics[width=\linewidth]{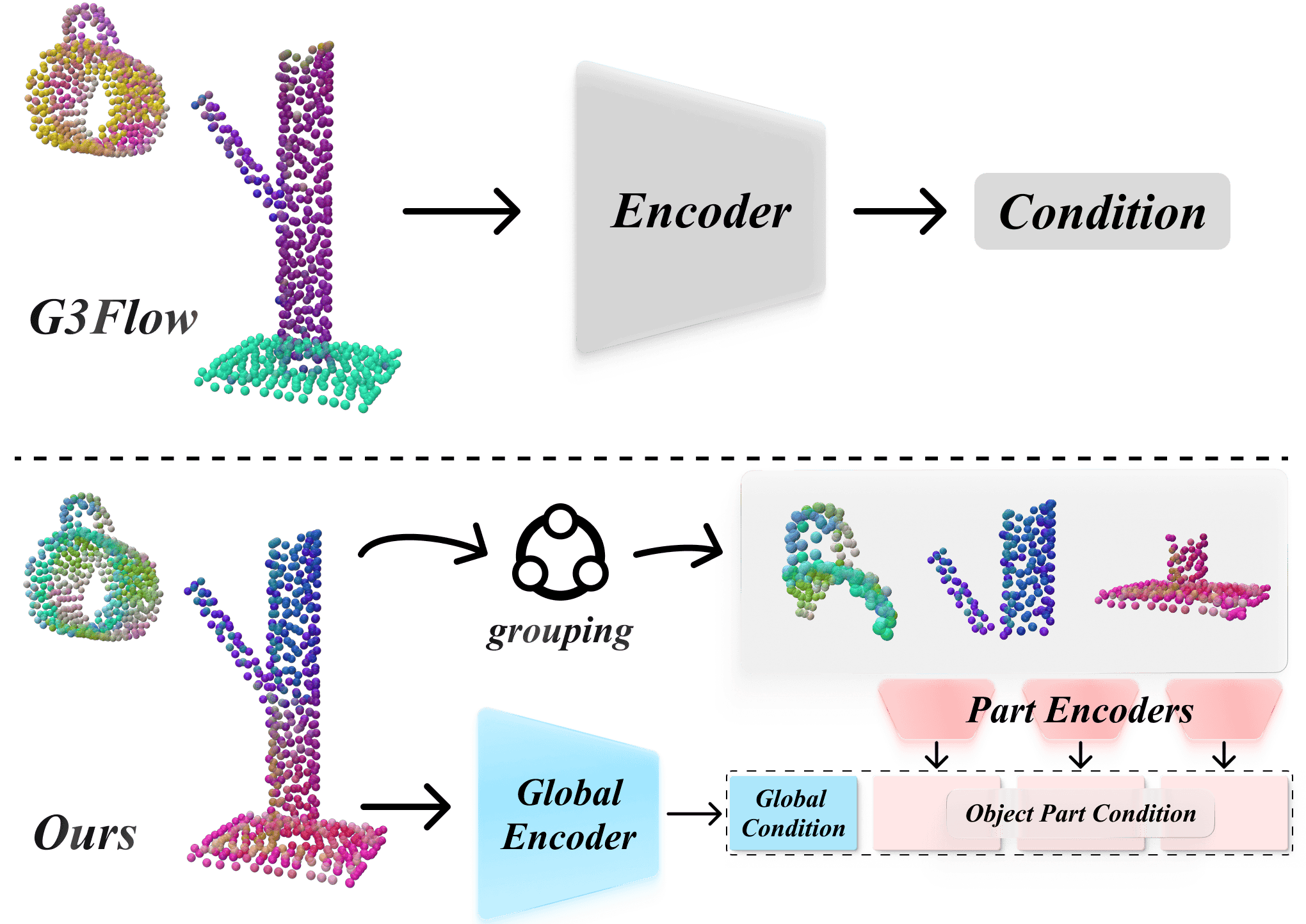}
   \caption{\textbf{Comparison of Conditioning Mechanisms.} 
    \textbf{Top:} The baseline employs a holistic conditioning approach, encoding the entire object point cloud into a single global vector, which lacks part-level details.
    \textbf{Bottom:} Our method uses a hierarchical approach. A global encoder captures overall context, while additional encoders extract complementary local features for fine-grained details. The resulting \textit{Hierarchical Condition} provides both global and local information, enabling more precise manipulation.}
   \vspace{-1.5em}
   \label{fig:motivation}
\end{figure}

To address it, recent work has focused on enriching 3D representations with semantic features~\cite{wang2023d,wang2024gendp}. Pioneering approaches like G3Flow~\cite{chen2025g3flow} construct a semantic field by leveraging powerful foundation models~\cite{oquab2023dinov2}, marking a step forward in semantic-aware manipulation.
However, despite its effectiveness, this method might yield a \textit{holistic} semantic representation (Fig.\ref{fig:motivation}~top), causing distinct part-level semantics to become indistinguishable. For example, as visualized in Fig.~\ref{fig:teaser}~(bottom left), features for a shoe's ``toe" and ``heel" become similar. Consequently, for pose-aware manipulation tasks that depend on differentiating these parts, the policy struggles to achieve the required precision.

In this work, we present \textbf{HeRO} (\textbf{H}ierarchical S\textbf{e}mantic \textbf{R}epresentation for \textbf{O}bject
manipulation) for part-level semantic perception of objects. Our central motivation is that \textit{pose-aware manipulation requires dense, fine-grained representations with strong spatial semantic coherence}. Motivated by recent advances in dense correspondence from foundation models \cite{zhang2023tale, tang2023emergent, stracke2025cleandift}, we first propose Dense
Semantic Lifting module to construct a semantic field that is \emph{denser and more discriminative} than G3Flow (Fig.~\ref{fig:teaser}, bottom-left). Concretely, we fuse features from DINOv2 \cite{oquab2023dinov2}, which are discriminative and geometrically precise for sparse correspondences, with features from Stable Diffusion (SD) \cite{rombach2022high}, which yield smooth, globally coherent correspondences. This complementary fusion preserves geometric accuracy while enforcing semantic consistency, enabling robust part-level correspondences for manipulation.

Given object point clouds and RGB-D observations, we apply \textbf{Dense Semantic Lifting} to extract a global semantic field $\mathcal{F}^G$ and a set of local semantic fields $\mathcal{F}^L=\left\{\mathcal{F}^{L,1}, \ldots, \mathcal{F}^{L,K}\right\}$. Specifically, we first extract complementary 2D features from DINOv2 and Stable Diffusion, fuse them through learnable weights, then lift these fused features to 3D by projecting each point onto the image plane and sampling the corresponding features. For $\mathcal{F}^G$, we create a temporally consistent global semantic field that combines geometric precision with semantic understanding through pose estimation between different timesteps. For $\mathcal{F}^L$, we partition $\mathcal{F}^G$ into $K$ sub-parts using PCA-based grouping to obtain semantically coherent local features (Fig.~\ref{fig:motivation}, bottom).

The dense features $\mathcal{F}^G$ and $\mathcal{F}^L$ are then fed into a {Hierarchical Conditioning Module} (HCM), where they serve as conditions for the generative process. Unlike conventional conditioning that concatenates conditions with the denoiser features, thereby introducing an order-sensitive inductive bias, the set $\mathcal{F}^L$ is inherently unordered (e.g., $\mathcal{F}^{L, 1}$ may correspond to either ``toe" or ``heel" across different shoes), which can confuse learning. We therefore adopt a {permutation-invariant} conditioning scheme \cite{qi2017pointnet++, wang2025pi3}: the HCM performs cross-attention between $\mathcal{F}^L$ and the denoising features {without} positional embeddings. As a result, the HCM effectively leverages the extracted hierarchical features, applies conditioning, and feeds them to the policy network for pose-aware object manipulation.

Through extensive experiments, HeRO sets a new state-of-the-art (SOTA) in pose-aware robotic manipulation. As shown in Fig.~\ref{fig:teaser} (bottom right), it improves success rate by 12.3\% over the prior best method, G3Flow~\cite{chen2025g3flow}, on the challenging \textit{Place Dual Shoes} task, and achieves an average gain of 6.5\% across six challenging pose-aware benchmark tasks. Overall, our main contributions are as follows: 
\begin{itemize} 
    \item We present HeRO, a framework for part-level semantic perception that employs Dense Semantic Lifting to construct fine-grained 3D semantic fields by fusing complementary features from DINOv2 and Stable Diffusion, thereby preserving geometric precision while ensuring semantic coherence.
    \item We propose a Hierarchical Conditioning Module (HCM) for diffusion-based policies, which integrates global context and a set of permutation-invariant, part-aware features, overcoming the limitations of holistic global conditioning.
    \item We provide extensive validation in both simulation and the real world, demonstrating that our method establishes a new state-of-the-art on challenging pose-aware manipulation benchmarks. 
\end{itemize}
\section{RELATED WORK}

\subsection{Diffusion Models for Imitation Learning}

Imitation learning~\cite{goyal2024rvt, zhao2023learning, chang2025generalizable, xia2025cage, xie2024decomposing} enables robots to acquire skills from expert demonstrations. Diffusion-based visuomotor policies~\cite{chi2023diffusion, liu_rdt-1b_2025} generate actions from 2D observations, producing smooth and temporally consistent trajectories, while flow-matching methods~\cite{zhang2025flowpolicy, sheng2025mp1} further improve training stability. Extensions to second-order flows~\cite{nguyen2025second} incorporate acceleration and jerk for even smoother motion. Most existing work, however, focuses on simple end-effectors and low-DoF control, with only a few addressing dexterous manipulation~\cite{weng2024dexdiffuser, liang2024skilldiffuser, liang2025dexhanddiff}, where stability and precision remain challenging.

The major limitation is their reliance on 2D features, which cannot fully capture spatial relationships and object-level semantics required for pose-aware manipulation. Recent studies show that integrating vision foundation models helps by lifting 2D semantic features into 3D representations, providing richer geometric context and semantic grounding that improve manipulation accuracy and robustness.

\subsection{Vision Foundation Models}

Vision foundation models have recently advanced dense correspondence learning along two complementary lines. Self-supervised Vision Transformers such as DINOv2~\cite{oquab2023dinov2} provide discriminative features well suited for semantic matching, while generative diffusion models~\cite{dhariwal2021diffusion, podell_sdxl_2024, tang2023emergent, stracke2025cleandift} yield dense and spatially coherent correspondences. Prior work has shown these representations to be complementary~\cite{zhang2023tale}, yet their synergy has not been fully explored in robotic manipulation. We leverage this complementarity by fusing DINO- and SD-derived features to construct 3D semantic fields that are both geometrically precise and semantically consistent, enabling robust part-level reasoning for pose-aware manipulation.

\begin{figure*}[thpb]
  \centering
  \includegraphics[width=\linewidth]{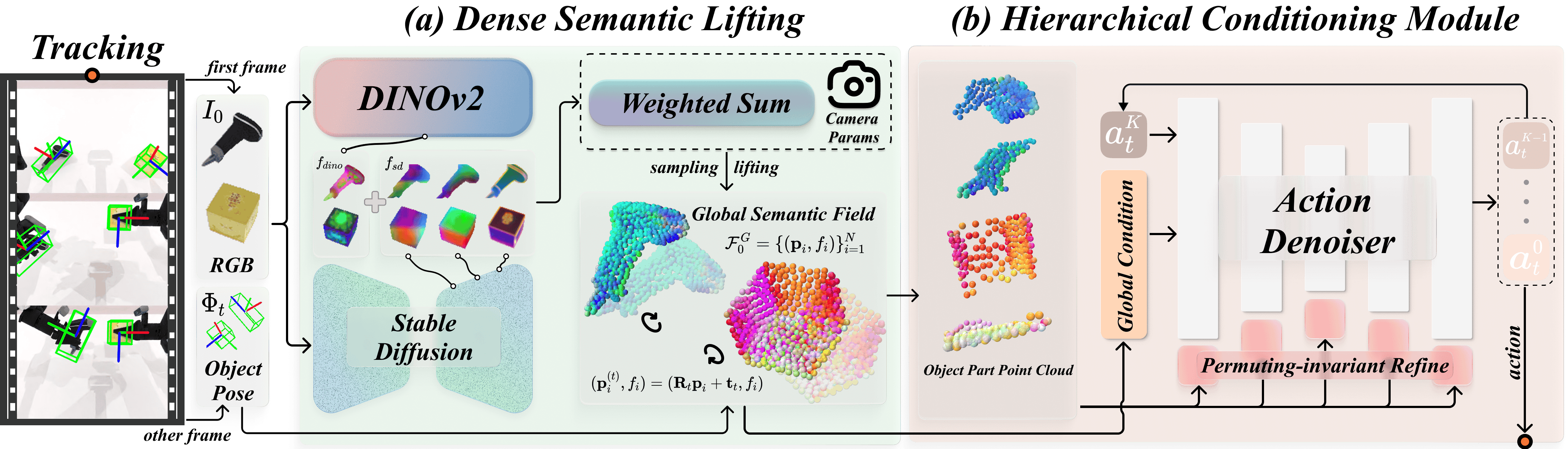}
  \caption{\textbf{Method Overview.} Our framework generates precise, pose-aware actions in a two-stage process.
  \textbf{(a) Dense Semantic Lifting:} We track object 6D poses from sequential frames and lift fused 2D features from DINOv2 (semantic) and Stable Diffusion (geometric) into an object-centric \textit{Dense Semantic Point Cloud}.
  \textbf{(b) Hierarchical Conditioning Module:} This point cloud is abstracted into a hierarchical \textit{Object Part Point Cloud}, which conditions the diffusion policy via two pathways: as a \textit{Global Condition} for the action denoiser and as fine-grained local guidance injected by a \textit{Permutation-invariant Refine} module at each denoising step. This dual mechanism enables precise, pose-aware action generation.}
  \label{fig:pipeline}
  \vspace{-1em}
\end{figure*}

\subsection{Semantic-Aware 3D Perception}

3D-based policies~\cite{ze20243d, ke_3d_2024, wilcox2025adapt3r} improve spatial reasoning in robotic manipulation by lifting two-dimensional features from pre-trained vision foundation models into 3D space, providing both geometric structure and semantic context. Some methods~\cite{wang2023d, wang2024gendp} create dense 3D descriptor fields from multi-view features, while others~\cite{ke_3d_2024, wilcox2025adapt3r} leverage lifted two-dimensional semantics to condition policies. Additional approaches~\cite{chen2025g3flow} generate continuous semantic flows to handle dynamic interactions and occlusions.

Despite these advances, most approaches emphasize global representations and lack fine-grained, part-level precision required for pose-aware manipulation. Our method addresses this by fusing DINOv2~\cite{oquab2023dinov2} and Stable Diffusion~\cite{rombach2022high} features to construct dense global and local semantic fields. This fusion preserves geometric accuracy while enforcing semantic consistency, enabling precise part-level correspondences for manipulation.

\section{METHOD}
Our framework enhances diffusion-based robotic manipulation through three key components, as illustrated in Fig.~\ref{fig:pipeline}.

First, we create rich semantic representations via \textbf{Dense Semantic Lifting} (\S\ref{sec:lifting}), which combines geometric point clouds with semantic features from visual foundation models to form Global and Local Semantic Fields. Second, we design a \textbf{Hierarchical Conditioning Module} (\S\ref{sec:conditioning}) that provides the diffusion policy with both global scene context and fine-grained part-level information through permutation-invariant conditioning. Finally, we train a \textbf{Diffusion Policy} (\S\ref{sec:policy}) that generates precise manipulation actions conditioned on these hierarchical semantic representations.

We begin by reconstructing 3D object geometry from multi-view RGB-D observations, following G3Flow~\cite{chen2025g3flow}, yielding dense point clouds $\mathcal{P}$. We downsample each point cloud using Farthest Point Sampling to $N=1024$ points: $\mathcal{P}_0 = \text{FPS}(\mathcal{P}, N)$.

\subsection{Dense Semantic Lifting}\label{sec:lifting}

While geometric point clouds provide spatial structure, effective manipulation requires understanding semantic properties like object parts, affordances, and material properties. Our Dense Semantic Lifting process enriches the geometric representation $\mathcal{P}_0$ with dense semantic features extracted from visual foundation models, creating a unified representation that combines both geometric precision and semantic understanding.

\noindent\textbf{Feature Extraction:} We leverage two complementary foundation models to capture different aspects of visual understanding for the first RGB frame $I_0$. DINOv2~\cite{oquab2023dinov2} provides discriminative, fine-grained visual features $f_{\text{dino}} \in \mathbb{R}^{H \times W \times d_v}$. In parallel, Stable Diffusion~\cite{rombach2022high} offers globally coherent semantic priors $f_{\text{sd}} \in \mathbb{R}^{H \times W \times d_s}$ by concatenating intermediate features from layers 2, 5, and 8, which encode rich semantic understanding developed through large-scale generative training. These complementary features capture both local visual details and global semantic context.

\noindent\textbf{Feature Fusion:} To unify these heterogeneous feature representations, we first apply PCA dimensionality reduction to project both feature maps to a common dimension $d$, yielding $f_{\text{dino}}' \in \mathbb{R}^{H \times W \times d}$ and $f_{\text{sd}}' \in \mathbb{R}^{H \times W \times d}$. The reduced features are then combined through a learnable weighted fusion:
\begin{equation}
f_{\text{fused}} = \alpha f_{\text{dino}}' + \beta f_{\text{sd}}',
\end{equation}
where $\alpha$ and $\beta$ are learnable parameters that adaptively balance the contribution.

\noindent\textbf{3D Lifting:} We transfer the fused 2D features to the 3D domain by projecting each point $\mathbf{p}_i \in \mathcal{P}_0$ onto the image plane using camera intrinsics and sampling the corresponding fused feature $f_i$ through bilinear interpolation:
\begin{equation}
\mathcal{F}_0^G = \{(\mathbf{p}_i^{(0)}, f_i)\}_{i=1}^N.
\end{equation}
This process creates the initial Global Semantic Field that combines precise geometric structure with rich semantic information, providing a foundation for downstream manipulation reasoning.

\noindent\textbf{Temporal Propagation:} As objects move during manipulation, we maintain temporal consistency of the semantic field by tracking each object's 6D pose trajectory $\Phi_t = (\mathbf{R}_t, \mathbf{t}_t) \in SE(3)$, at each timestep $t$. The semantic field is updated by applying rigid-body transformations to point positions while preserving their associated semantic features, which represent intrinsic object properties independent of pose:

\begin{equation}
(\mathbf{p}_i^{(t)}, f_i) = (\mathbf{R}_t \mathbf{p}_i^{(t)} + \mathbf{t}_t, f_i), \quad i=1,\dots,N.
\end{equation}
This transformation yields the updated Global Semantic Field at timestep $t$:
\begin{equation}
\mathcal{F}_t^G = \{(\mathbf{p}_i^{(t)}, f_i)\}_{i=1}^N.
\end{equation}
Through this temporal propagation strategy, we construct the complete Global Semantic Field $\mathcal{F}_G = \{\mathcal{F}_t^G\}_{t=0}^T$ that maintains both geometric and semantic consistency across the entire manipulation sequence.

\subsection{Hierarchical Conditioning Module}\label{sec:conditioning}
\begin{figure}[tbp]
   \centering
   \includegraphics[width=\linewidth]{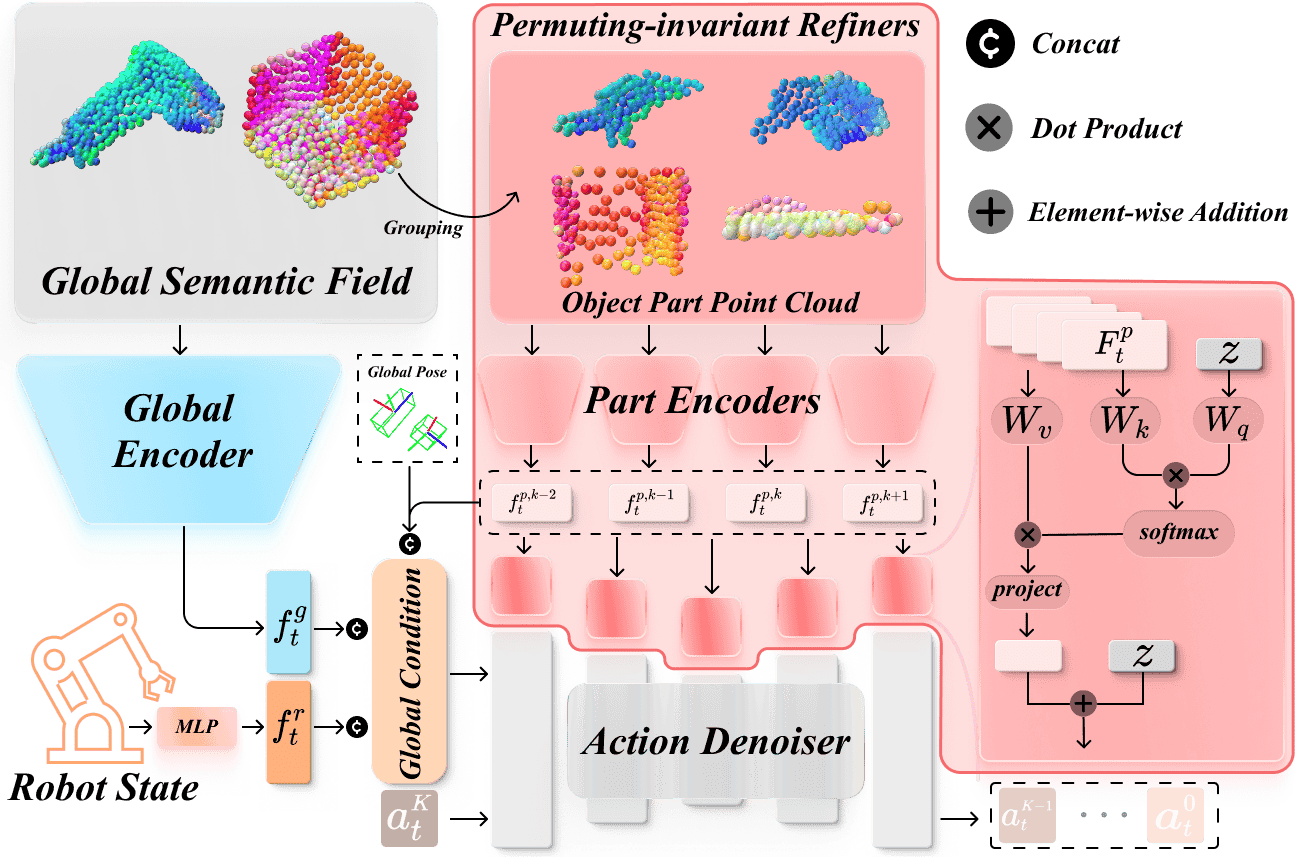}
   \caption{\textbf{Hierarchical Conditioning Module Architecture.} Our model uses a dual-pathway design to guide the Action Denoiser. The \textbf{Global Path} processes the entire point cloud into a single global condition for high-level context. The \textbf{Local Path} partitions the point cloud into semantic parts, which are encoded by Permutation-invariant Refiners into a set of fine-grained embeddings. These local conditions are injected into the denoiser, enabling actions that are both globally consistent and locally precise.}
   \label{fig:network}
   \vspace{-1em}
\end{figure}

While $\mathcal{F}_G$ provides comprehensive scene understanding, effective manipulation requires fine-grained reasoning about individual object parts and their relationships. For instance, grasping different parts of a shoe (heel vs. toe) requires distinct manipulation strategies. We address this by conditioning the diffusion policy through dual pathways: global scene context and part-level information, as illustrated in Fig.~\ref{fig:network}.

\noindent\textbf{Local Field Construction:} At each timestep $t$, we decompose $\mathcal{F}_t^G$ into Local Semantic Fields that capture semantically meaningful object parts:
$
\mathcal{F}_{t}^L = \{\mathcal{F}_{t}^{L,k}\}_{k=1}^{K}
$
where $K=8$ represents the number of part-level clusters. We create an augmented representation for each point by concatenating spatial coordinates with semantic features:
$
\mathbf{x}_i = [\mathbf{p}_i^{(t)}; f_i] \in \mathbb{R}^{3+d}
$
where $\mathbf{p}_i^{(t)} \in \mathbb{R}^3$ captures geometric structure and $f_i \in \mathbb{R}^d$ encodes semantic properties.

We apply PCA to identify the dominant structural variation across all points. Points are sorted along the first principal component, which typically aligns with the object's main elongation axis, and evenly partitioned into $K$ clusters. This PCA-based approach naturally discovers part boundaries while ensuring each local field $\mathcal{F}_{t}^{L,k}$ represents a spatially and semantically coherent object region.

\subsubsection{Global Conditioning}
For effective diffusion-based action generation, the policy requires comprehensive contextual information integrating environmental understanding and robot state awareness. We construct this context through three complementary feature streams:

\noindent\textbf{Scene Features:} We encode the global semantic field $\mathcal{F}_t^G$ using a PointNet encoder to extract scene feature $f_t^g$. This captures overall geometric structure, semantic context, and spatial relationships between objects in the scene.

\noindent\textbf{Robot Features:} We encode the robot's joint states through a multi-layer perceptron (MLP) to produce robot feature $f_t^r$. This provides information about the manipulator's current configuration, pose constraints, and kinematic limitations that influence action feasibility.

\noindent\textbf{Part Features:} We individually encode each local field $\mathcal{F}_{t}^{L,k}$ using PointNet encoders and aggregate them using current object poses $\Phi_t$ to generate part feature $f_t^p$. This preserves fine-grained part-level distinctions while incorporating object-specific geometric context.

The three feature streams are concatenated to form a comprehensive global conditioning vector:
\begin{equation}
f_t^{\text{global}} = \text{Concat}\big(f_t^g, f_t^r, f_t^p\big).
\end{equation}
This unified representation integrates scene-level semantics, robot kinematics, and part-level object understanding, providing contextual information for informed action generation.

\subsubsection{Permutation-Invariant Part Conditioning}
The local fields $\mathcal{F}_{t}^L$ represent unordered collections of object parts where the assignment of parts to indices varies unpredictably across different objects (e.g., $\mathcal{F}_{t}^{L,1}$ could be either heel or toe for different shoes). Conventional conditioning approaches that rely on concatenation or positional encodings would introduce order-sensitive biases. To address this challenge, we design a permutation-invariant conditioning pipeline that processes part features through attention mechanisms {without positional embeddings}, as shown in the Fig.~\ref{fig:network}~(right).

\noindent\textbf{Part Feature Extraction:} We encode each local field $\mathcal{F}_{t}^{L,k}$ using PointNet to extract part-specific features:
\[
F_t^p = \{f_{t}^{p,k}\}_{k=1}^{K}.
\]

\noindent\textbf{Inter-Part Reasoning:} We apply self-attention {without positional embeddings} to enable information exchange between different object parts while preserving permutation invariance across part orderings:
\begin{equation}
\hat{F}_t^p = \mathcal{A}_{\text{self}}(F_t^p).
\end{equation}
The absence of positional embeddings ensures that the attention mechanism remains invariant to arbitrary permutations of the part set.

\noindent\textbf{Cross-Attention Conditioning:} The refined part features are injected into the diffusion U-Net through cross-attention layers, where U-Net features serve as queries and part features serve as keys and values:
\begin{equation}
z_{\text{next}} = z + \mathcal{A}_{\text{cross}}(z, \hat{F}_t^p).
\end{equation}
This cross-attention mechanism allows the diffusion model to condition on detailed part-level information while maintaining permutation invariance.

This three-stage pipeline enables robust part-level reasoning without sensitivity to arbitrary part orderings, ensuring consistent performance across diverse object configurations.

\subsection{Diffusion Policy Learning}\label{sec:policy}

We train a diffusion model to generate robot actions. Given expert actions $A_i^*$, the model learns to predict noise $\epsilon$ added at diffusion timestep $t$.

The network $\epsilon_\theta$ uses all extracted features as conditions. The loss is:
\begin{equation}
    \mathcal{L} = \mathbb{E}_{A_i^*, c_i, \epsilon, t} \Big[ \big\| \epsilon - \epsilon_\theta(a_t, t, c_i) \big\|^2 \Big],
\end{equation}
where $a_t$ is the noisy action. During inference, we iteratively denoise random Gaussian noise to generate actions conditioned on the extracted features.

\section{EXPERIMENTS}
We conduct a series of experiments to validate our proposed method. Through quantitative benchmarks and qualitative analysis, we aim to answer the following questions:

(1) \textbf{Semantic Requirements:} How critical is semantic understanding for pose-aware manipulation tasks?

(2) \textbf{Semantic Representation Quality:} Why does our semantic representation outperform G3Flow?

(3) \textbf{Fine-grained Object Perception:} What is the specific contribution of fine-grained geometric perception to policy performance in pose-sensitive scenarios?

\subsection{Experimental Setup}

\begin{figure}[tbp]
   \centering
   \includegraphics[width=0.8\linewidth]{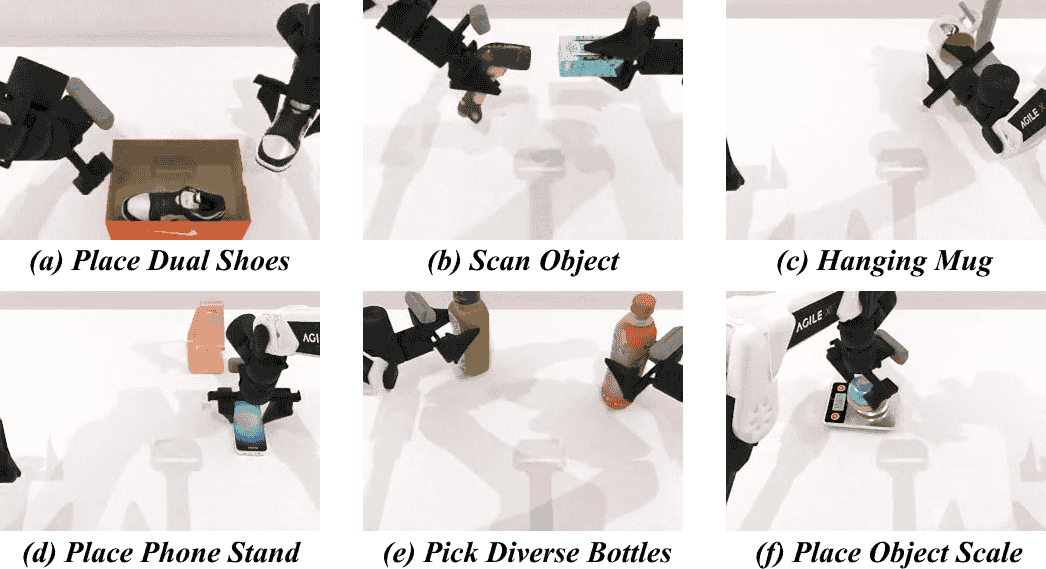}
   \caption{\textbf{Simulated Manipulation Tasks.} We evaluate our method on six challenging tasks from the RoboTwin 2.0 benchmark. These tasks necessitate precise pose estimation and a nuanced understanding of object-part semantics to facilitate successful interaction.}
   \label{fig:simulated_experiments_tasks}
   \vspace{-0.5em}
\end{figure}
\begin{figure}[tbp]
   \centering
   \includegraphics[width=0.8\linewidth]{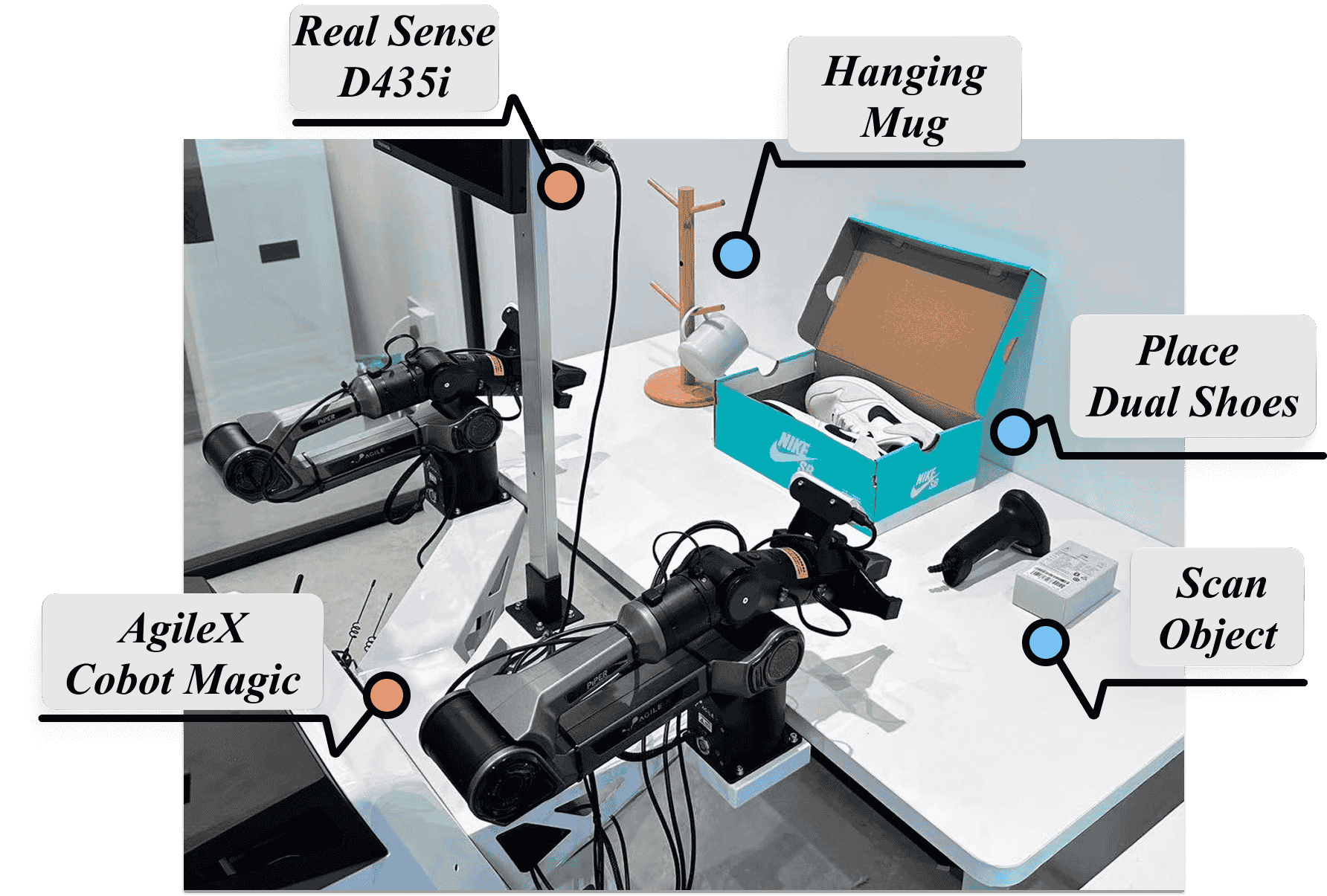}
   \caption{\textbf{Real-World Experimental Setup.} Our experimental setup consists of an AgileX Cobot Magic dual-arm robot equipped with a RealSense D435i head-mounted camera.}
   \label{fig:real_world_setup}
   \vspace{-1em}
\end{figure}

\noindent\textbf{Tasks.} We build upon the multi-object tasks(Place Dual Shoes, Pick Diverse Bottles) from G3Flow~\cite{chen2025g3flow} and further select several challenging tasks from RoboTwin 2.0~\cite{mu2025robotwin} that require precise object pose alignment or involve complex object interactions, as shown in Fig.~\ref{fig:simulated_experiments_tasks}. We also selected three of these tasks for cross-object generalization experiments, real-world experiments validation, and ablation.

\noindent\textbf{Baselines.} We compare our method with several state-of-the-art approaches to validate the effectiveness:
\textbf{G3Flow}~\cite{chen2025g3flow} is a foundation model-driven approach that constructs semantic flow by integrating 3D generative models, vision foundation models, and pose tracking, yielding dynamic object-centric representations. \textbf{Diffusion Policy (DP)}~\cite{chi2023diffusion} formulates visuomotor policy learning as a conditional denoising diffusion process, predicting actions from image inputs for robotic manipulation. Building on this, \textbf{3D Diffusion Policy (DP3)}~\cite{ze20243d} introduces a lightweight MLP encoder for sparse point clouds to enable efficient 3D representation learning, with a variant (\textbf{DP3 w/ color}) that further incorporates RGB features projected onto the point clouds.

\newcommand{ \rbf}[1]{\text{\scriptsize\textcolor{gray}{\ensuremath{\pm#1}}}}
\begin{table*}[t]
\small
\centering
\caption{\textbf{Comparative Evaluation on the 6 tasks in RoboTwin 2.0 Benchmark with standard setting.}}
\vspace{-0.5em}
\resizebox{\textwidth}{!}{
\begin{tabular}{l|cccccc|c}
\toprule
\textbf{Method} & 
\textbf{\textit{\shortstack{Place Dual\\Shoes}}} & 
\textbf{\textit{\shortstack{Scan\\Object}}} & 
\textbf{\textit{\shortstack{Hanging\\Mug}}} & 
\textbf{\textit{\shortstack{Place Phone\\Stand}}} & 
\textbf{\textit{\shortstack{Pick Diverse\\Bottles}}} & 
\textbf{\textit{\shortstack{Place Object\\Scale}}} & 
\textbf{\textit{Average}} \\ 
\midrule
DP~\cite{chi2023diffusion} 
& 7.0 \rbf{3.6} 
& 10.3 \rbf{2.1} 
& 15.3 \rbf{2.6} 
& 37.7 \rbf{6.2} 
& 18.3 \rbf{3.1} 
& 2.7 \rbf{1.2} 
& 15.2 \\

DP3~\cite{ze20243d} 
& 17.7 \rbf{3.4} 
& 23.3 \rbf{1.2} 
& 23.3 \rbf{6.3} 
& 52.0 \rbf{2.9} 
& 27.7 \rbf{7.0} 
& 10.7 \rbf{3.9} 
& 25.8 \\

DP3 w/ color~\cite{ze20243d} 
& 16.3 \rbf{2.9} 
& 14.6 \rbf{1.2} 
& 21.7 \rbf{5.6} 
& 41.7 \rbf{6.1} 
& 17.0 \rbf{5.1} 
& 4.7 \rbf{1.2} 
& 19.3 \\

G3Flow~\cite{chen2025g3flow} 
& 20.7 \rbf{3.4} 
& 22.0 \rbf{3.7} 
& 26.7 \rbf{2.5} 
& 45.3 \rbf{3.3} 
& 32.0 \rbf{8.0} 
& 7.7 \rbf{0.9} 
& 25.7 \\ 
\midrule

\textbf{Ours} 
& \best{33.0 \rbf{5.7}}
& \best{26.7 \rbf{0.5}}
& \best{31.0 \rbf{1.4}} 
& \best{55.3 \rbf{1.9}} 
& \best{34.7 \rbf{3.8}}
& \best{11.7 \rbf{1.2}} 
& \best{32.3} \\

\bottomrule
\end{tabular}}
\vspace{-1em}
\label{tab:sota_standard}%
\end{table*}

\noindent\textbf{Training and Evaluation Details.} 
For fair comparison, we reproduce G3Flow and retrain all baseline methods in the official RoboTwin 2.0 benchmark codebase. Following best practices from prior work rather than the default RoboTwin 2.0 leaderboard setting (50 demonstrations), we use 100 expert demonstrations for training in simulation enviorment. 

All methods are trained with their recommended hyperparameters: 3000 epochs with batch size 256 for our method, DP3 and G3Flow, and 600 epochs with batch size 128 for DP. For standard benchmark tasks, we use all official assets for both training and evaluation. For cross-object generalization, we train on two-thirds of the provided assets and evaluate on the remaining one-third. Following G3Flow~\cite{chen2025g3flow}, we use GroundedSAM~\cite{ren2024grounded} for object segmentation and FoundationPose~\cite{wen2024foundationpose} for 6D object tracking. We extract and fuse features from DINOv2~\cite{oquab2023dinov2} and finetuned Stable Diffusion~\cite{stracke2025cleandift} to construct the semantic point cloud.
For real-world experiments, we use an AgileX Cobot Magic dual-arm embodiment equipped with RealSense D435i cameras as illustrated in Fig.~\ref{fig:real_world_setup}. We collect 50 expert demonstrations for each task using teleoperation.

\noindent\textbf{Evaluation Protocol.} 
To ensure fair evaluation, we fix random seeds across training and testing. Each method is run with multiple seeds and evaluated on 100 test episodes per seed, reporting mean success rates and standard deviations. In simulation, object positions and poses are randomized following the benchmark settings. For real-world evaluation, we conduct 20 episodes with fixed object positions but randomized rotations. All Inference runs on a single 4090D.

\subsection{Comparision of the State-of-the-Art}
We conduct three distinct experiments against state-of-the-art baselines: a standard benchmark evaluation, a cross-object generalization test and real-world validation.

\noindent\textbf{Standard Benchmark Performance.}
We first evaluation assesses performance under a standard benchmark protocol, wherein all methods are trained and tested on the same set of object assets. This "closed-set" configuration measures a policy's ability to master tasks within a known environment. As presented in Table~\ref{tab:sota_standard}, our method establishes a new state-of-the-art, achieving an impressive average success rate of 32.3\%. This result represents a significant 6.6\% improvement over G3Flow, the strongest baseline. The superiority of our approach becomes particularly evident in tasks demanding precise semantic alignment. For instance, in \textit{Place Dual Shoes} and \textit{Hanging Mugs}, our method surpasses G3Flow by a substantial margin of 12.3\% and 4.3\%, respectively. This large performance gap highlights the efficacy of our hierarchical semantic conditioning, which excels at capturing and leveraging fine-grained, part-level object details. In contrast, while DP3-based methods show reasonable performance on geometrically simpler tasks like \textit{Place Phone Stand}, their success rates decline sharply on more complex, pose-sensitive scenarios. This outcome confirms our hypothesis that relying solely on 3D geometric information is insufficient; explicit semantic reasoning is indispensable for achieving robust and precise manipulation.

\begin{table}[t]
\small
\centering
\caption{\textbf{Cross-object generalization results (success rates in \%).}}
\vspace{-0.5em}
\resizebox{\linewidth}{!}{
\begin{tabular}{l|ccc|c}
\toprule
\textbf{Method} & 
\textbf{\textit{\shortstack{Place Dual\\Shoes}}} & 
\textbf{\textit{\shortstack{Scan\\Object}}} & 
\textbf{\textit{\shortstack{Hanging\\Mug}}} & 

\textbf{\textit{Average}} \\
\midrule
DP~\cite{chi2023diffusion} & 
7.3 \rbf{2.1} & 
4.0 \rbf{2.2} & 
5.0 \rbf{2.2} & 
5.4 \\

DP3~\cite{ze20243d} & 
11.3 \rbf{3.4} & 
11.3 \rbf{1.2} & 
22.7 \rbf{1.9} & 
15.1 \\

DP3 w/ color~\cite{ze20243d} & 
10.0 \rbf{1.4} & 
8.3 \rbf{1.2}& 
19.7 \rbf{4.2} & 
12.7 \\

G3Flow~\cite{chen2025g3flow} & 
16.3 \rbf{2.9} & 
11.7 \rbf{2.6} & 
25.0 \rbf{3.7} & 
17.7 \\

\midrule
\textbf{Ours} & 
\best{28.3} \rbf{5.3} & 
\best{14.3} \rbf{0.5}& 
\best{30.7} \rbf{4.2} & 
\best{24.4} \\

\bottomrule
\end{tabular}}
\label{tab:sota_generalize}
\vspace{-0.5em}
\end{table}

\begin{table}[t]
\small
\centering
\caption{\textbf{Real-world experiment results (success rates in \%).}}
\vspace{-0.5em}
\begin{tabular}{l|ccc|c}
\toprule
\textbf{Method} & 
\textbf{\textit{\shortstack{Place Dual\\Shoes}}} & 
\textbf{\textit{\shortstack{Scan\\Object}}} & 
\textbf{\textit{\shortstack{Hanging\\Mug}}} & 
\textbf{\textit{Average}} \\
\midrule
DP3~\cite{ze20243d} 
& 5 
& 15 
& 0 
& 6.7 \\

G3Flow~\cite{chen2025g3flow} 
& 10 
& 30 
& 10 
& 16.7 
\\ 

\midrule

\textbf{Ours} 
& \best{25} 
& \best{35} 
& \best{20} 
& \best{26.7} \\

\bottomrule
\end{tabular}
\label{tab:real_world_results}
\vspace{-0.5em}
\end{table}

\noindent\textbf{Generalization to Unseen Objects.}
More rigorous evaluation specifically tests zero-shot generalization to novel objects. In this "open-set" protocol, policies are evaluated on object instances that were explicitly held out from the training set. This setup is a critical test of a model's ability to abstract and transfer knowledge beyond mere memorization. As shown in Table~\ref{tab:sota_generalize}, our method again demonstrates superior capabilities, consistently outperforming all baselines with an average success rate of 24.4\%, which is a notable 6.7\% lead over G3Flow. This success stems from our hierarchical semantic representation, which learns to abstract functional and geometric properties rather than overfitting to the specific visual characteristics of the training instances. For example, in the challenging \textit{Place Dual Shoes} task, our method achieves a 12\% higher success rate than G3Flow. This result underscores its ability to comprehend and align abstract semantic parts, such as the toe of one shoe to the heel of another, even on previously unseen models. Conversely, the performance of all baseline methods deteriorates significantly in this setting. Their inability to handle variations in object geometry and appearance highlights a critical reliance on memorizing the training data, rather than acquiring a truly generalizable manipulation skill.
\begin{figure*}[thpb]
  \centering
  \includegraphics[width=0.9\linewidth]{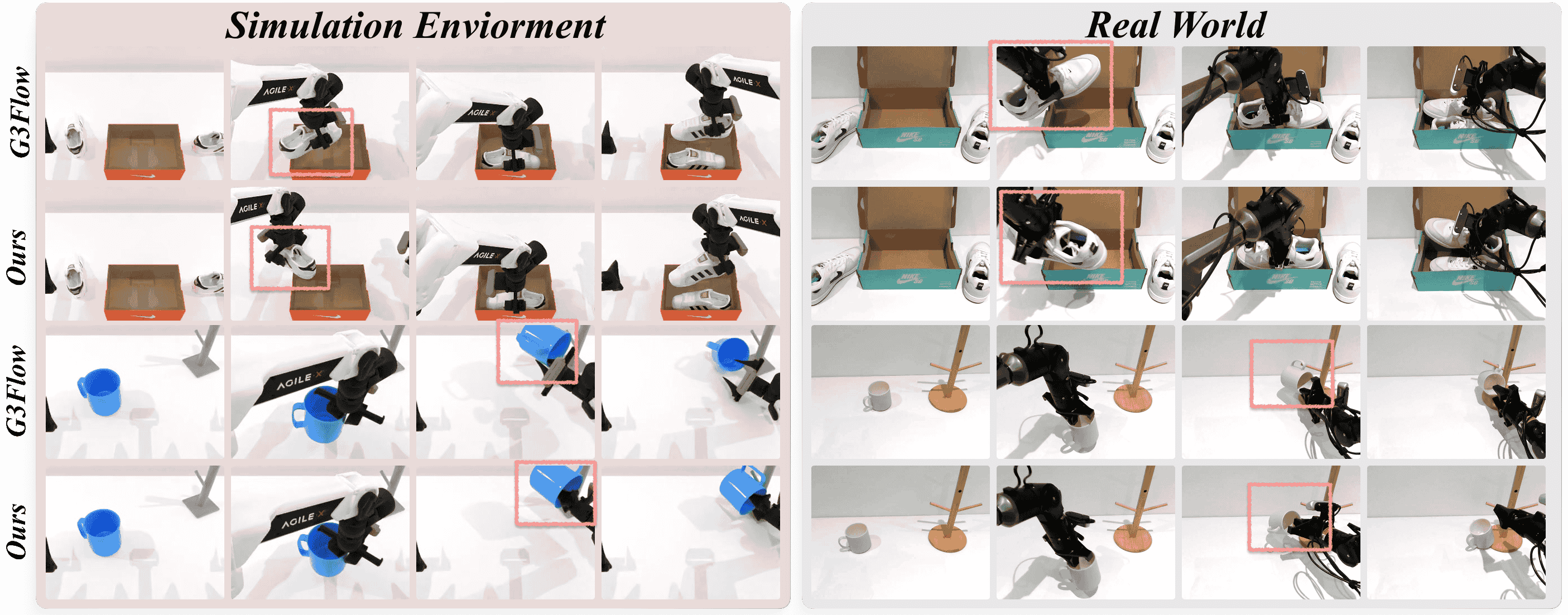}
  \caption{\textbf{Qualitative comparison of policy execution.} For \textit{Place Dual Shoes} (left), G3Flow executes an erroneous rotation, resulting in task failure, whereas our method achieves correct alignment. For \textit{Hanging Mug} (right), G3Flow fails to align the handle with the rack. In contrast, our policy's fine-grained semantic representation enables precise handle orientation and successful task completion, both in simulated environments and on the physical robot.}
  \label{fig:task_visualization}
  \vspace{-1em}
\end{figure*}

\noindent\textbf{Real-World Validation.}
We also conducted experiments entirely in the real-world setup where policies were trained using data collected directly from our robotic embodiment via teleoperation, and subsequently evaluated on the same hardware. 
As detailed in Table~\ref{tab:real_world_results}, our method demonstrates strong and reliable performance, achieving the highest success rates across all evaluated tasks. This outcome is particularly significant as it validates that our hierarchical semantic representation is not only effective in theory but also robust enough to handle the nuances of a non-simulated environment. The superior performance compared to the baselines underscores our model's ability to learn meaningful and actionable policies from real-world data.

\begin{figure}[tbp]
   \centering
   \includegraphics[width=\linewidth]{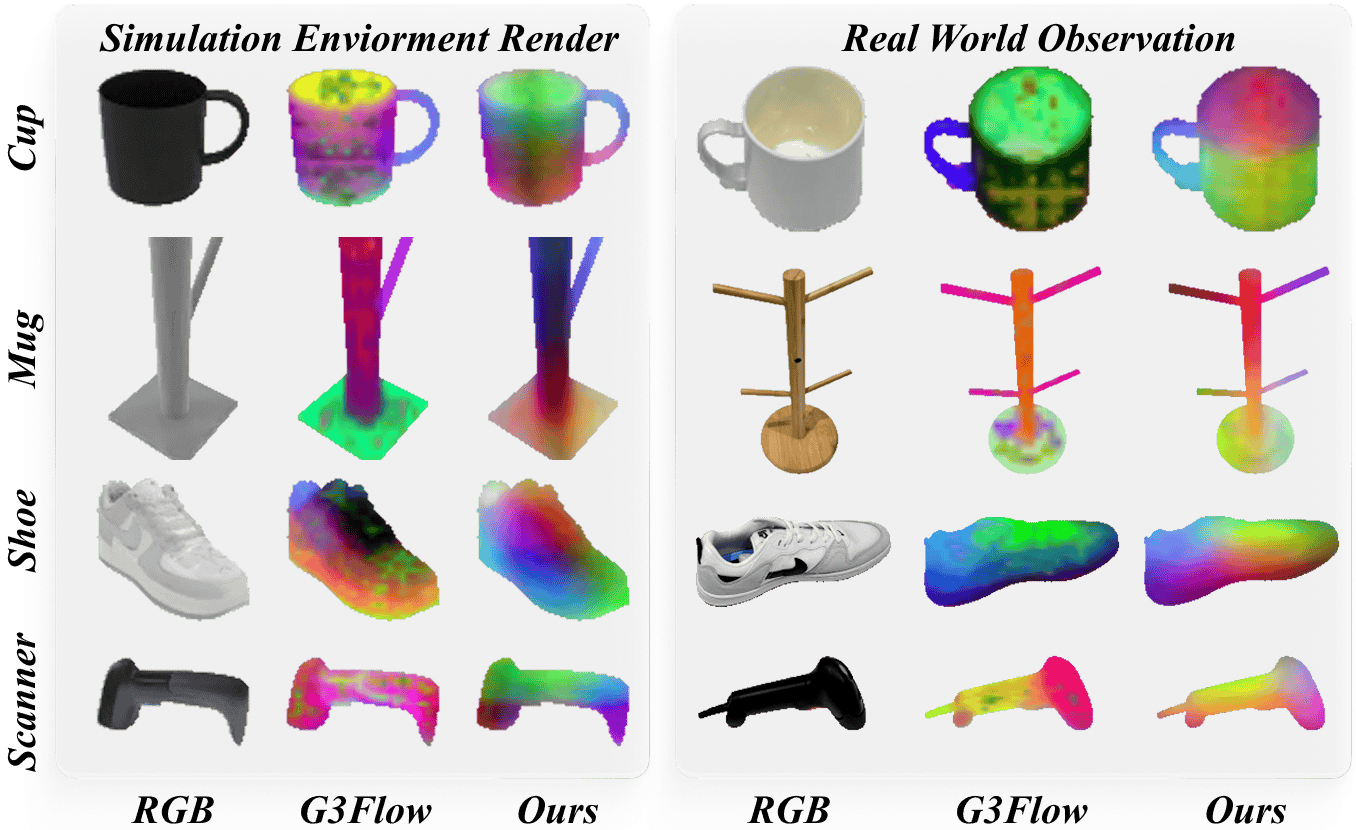}
   \caption{\textbf{Visualization of object semantics field.} We visualize the semantics field of G3Flow and our method in simulation versus real-world data. G3Flow exhibit significant noise and geometric inconsistency. In contrast, our method generates smooth, coherent semantic fields that robustly delineate functional parts across both domains, providing a more stable representation for the policy.}
   \label{fig:semantics_visualization}
   \vspace{-2em}
\end{figure}

\subsection{Analysis with visualization.}
To address our second research question regarding the superior quality of our semantic representation, we conduct a two-part qualitative analysis. We first compare the execution of our policy against G3Flow on representative tasks, then delve into a detailed examination of the underlying semantic feature fields to diagnose the performance gap.

\noindent\textbf{Qualitative Policy Performance.} We first visualize policy rollouts on challenging, pose-aware tasks in Fig.~\ref{fig:task_visualization}. The comparison reveals G3Flow's critical limitations, which stem from its inability to explicitly recognize part-level semantics. For instance, in the \textit{Place Dual Shoes} task, G3Flow's holistic representation fails to distinguish between the shoe's toe and heel, leading to an erroneous rotation and an incorrect final placement that results in task failure. Similarly, in the \textit{Hanging Mug} task, G3Flow is unable to isolate the handle as a key functional part, causing the policy to grasp the mug's body and fail to securely hook it onto the rack. In stark contrast, our policy consistently succeeds by leveraging a fine-grained understanding of object geometry and semantics. It correctly identifies the distinct parts of the shoe for precise alignment and isolates the mug handle for a successful grasp and hang, demonstrating a significantly enhanced capability for precise, real-world manipulation.

\noindent\textbf{Analysis of Semantic Representation.} To further diagnose the performance gap, we analyze the underlying semantic representations that drive policy decisions in Fig.~\ref{fig:semantics_visualization}. The quality of these learned dense object semantic field is crucial, as they form the basis for the policy's understanding of object structure and orientation. The visualizations reveal that G3Flow's semantic field are often noisy and inconsistent. The color gradients, representing semantic features, appear fragmented and do not align coherently with the object's underlying geometry, indicating a confused representation. This instability is exacerbated in real-world conditions, where lighting and texture variations lead to further degradation. Our method, however, produces significantly smoother and more geometrically consistent semantics. The color gradients transition logically across surfaces, clearly delineating functional parts like the cup handle or shoe toe with uniform and distinct feature representations, both in rendered simulation objects and real-world images. This stability and accuracy provide our policy with a reliable and unambiguous foundation for executing precise manipulation, directly explaining its superior performance.

\subsection{Ablation Study}
\begin{table}[tbp]
    \centering
    \caption{\textbf{Ablation Study of Model Components.} We evaluate the contribution of each key module by measuring the average success rate (\%) across 3 benchmark tasks.}
    \label{tab:ablation}
    \vspace{-0.5em}
    \begin{tabular}{ccc|c}
        \toprule
        \makecell{\textbf{Dense}\\\textbf{Semantic}} & 
        \makecell{\textbf{Global Pose-}\\\textbf{aware Condition}} & 
        \makecell{\textbf{Part-aware}\\\textbf{Geometry Refine}} & 
        \textbf{Average (\%)} \\
        \midrule
        & & & 23.1 \\
        \checkmark & & & 23.7\\
         & \checkmark & & 24.3 \\
         & & \checkmark & 27.6 \\
        \checkmark & \checkmark & \checkmark & \best{30.2} \\
        \bottomrule
    \end{tabular}
    \vspace{-2em}
\end{table}
To answer our third research question regarding the importance of fine-grained perception, we conducted an ablation study as shown in Table~\ref{tab:ablation}. 
The results clearly indicate that the \textit{Part-aware Geometry Refinement} module is the most critical component. Adding it to the baseline model yields the largest performance gain, from 23.1\% to 27.6\%. This directly validates that a fine-grained understanding of object geometry is essential for precise, pose-aware manipulation. While other components like semantic features and pose conditioning provide benefits, it is the combination of all modules that achieves the highest success rate of 30.2\%, demonstrating a powerful synergy between them.

\section{CONCLUSIONS}
We introduced \textbf{HeRO}, a framework for part-level semantic perception in pose-aware object manipulation. By fusing discriminative DINOv2 features with globally coherent Stable Diffusion features, HeRO constructs dense 3D semantic fields that maintain both geometric precision and semantic consistency. The Hierarchical Conditioning Module effectively integrates global context with permutation-invariant, part-aware features, enabling diffusion-based policies to leverage structured hierarchical information for precise manipulation. Extensive experiments demonstrate HeRO outperforms prior methods, achieving state-of-the-art performance.                                

\section*{ACKNOWLEDGMENT}

This work was supported in part by National Natural Science Foundation of China under grant No.62372091 and in part by Hainan Province Science and Technology SpecialFund under grant No. ZDYF2024(LALH)001.
\bibliography{IEEEabrv}
\bibliographystyle{IEEEtran}

\end{document}